\documentclass[letterpaper, 10 pt, conference]{ieeeconf}  

\IEEEoverridecommandlockouts                              

\makeatletter
\newcommand*\titleheader[1]{\gdef\@titleheader{#1}}
\AtBeginDocument{%
  \let\st@red@title\@title
  \def\@title{\vspace{-2em}%
    \bgroup\normalfont\large\centering\@titleheader\par\egroup
    \vskip1.5em\st@red@title}
}
\makeatother

\usepackage{graphicx} 
\usepackage{multirow}
\usepackage{array}
\usepackage{amsmath} 
\usepackage{url} 
\usepackage{hyperref}
\usepackage{amsfonts} 
\usepackage[final]{changes}

\usepackage{booktabs}
\usepackage{color, colortbl}
\definecolor{Gray}{gray}{0.95}
\titleheader{2024 IEEE/RSJ International Conference on Robots and Systems (IROS)}

\title{\LARGE \bf
Bridging Language, Vision and Action: Multimodal VAEs in Robotic Manipulation Tasks
}

\author{Gabriela Sejnova$^{1}$, Michal Vavrecka$^{1}$, Karla Stepanova$^{1}$
\thanks{}
\thanks{$^{1}$ Czech Institute of Informatics, Robotics and Cybernetics, Czech Technical University in Prague, Prague, Czech Republic,
        {\tt\small [gabriela.sejnova, karla.stepanova]@cvut.cz}}%
}

\begin{document}

\maketitle
\thispagestyle{empty}
\pagestyle{empty}

\begin{abstract}

In this work, we focus on unsupervised vision-language-action mapping in the area of robotic manipulation.
Recently, multiple approaches employing pre-trained large language and vision models have been proposed for this task. However, they are computationally demanding and require careful fine-tuning of the produced output. A more lightweight alternative would be the implementation of multimodal Variational Autoencoders (VAEs) which can extract the latent features of the data and integrate them into a joint representation, as has been demonstrated mostly on image-image or image-text data for the state-of-the-art models. Here, we explore whether and how multimodal VAEs can be employed in unsupervised robotic manipulation tasks in a simulated environment. Based on the results obtained, we propose a model-invariant training alternative that improves the models' performance in a simulator by up to 55~\%. 
 Moreover, we systematically evaluate the challenges raised by individual tasks, such as object or robot position variability, number of distractors, or task length. Our work thus also sheds light on the potential benefits and limitations of using the current multimodal VAEs for unsupervised learning of robotic motion trajectories based on vision and language. 


\end{abstract}

\section{INTRODUCTION}

Teaching robots new manipulation tasks by demonstration/kinesthetic guidance is one of the key research areas in robotics. Here we focus on the scenario where the robot also receives a visual observation of the scene and natural language (NL) description of the task and has to fuse this information with kinesthetic teaching to successfully fulfill the goal. While traditional rule-based approaches usually parse the NL input into a formal language for task disambiguation \cite{behrens2019specifying}, \cite{tellex2020robots}, the most recent methods utilize large language (LLM) \cite{shah2023lm}, \cite{liang2023code}, \cite{wu2023tidybot} and large vision models (LVM) \cite{zitkovich2023rt}, \cite{driess2023palm} pre-trained on internet-scale data, which are additionally fine-tuned, e.g., via prompting methods. However, these models often \deleted{generate biased outputs,} struggle with repeatability and can be more difficult to control in terms of predictable outputs \cite{chang2023survey}.

\begin{figure}[!htp]
\centering
\includegraphics[width=1\columnwidth]{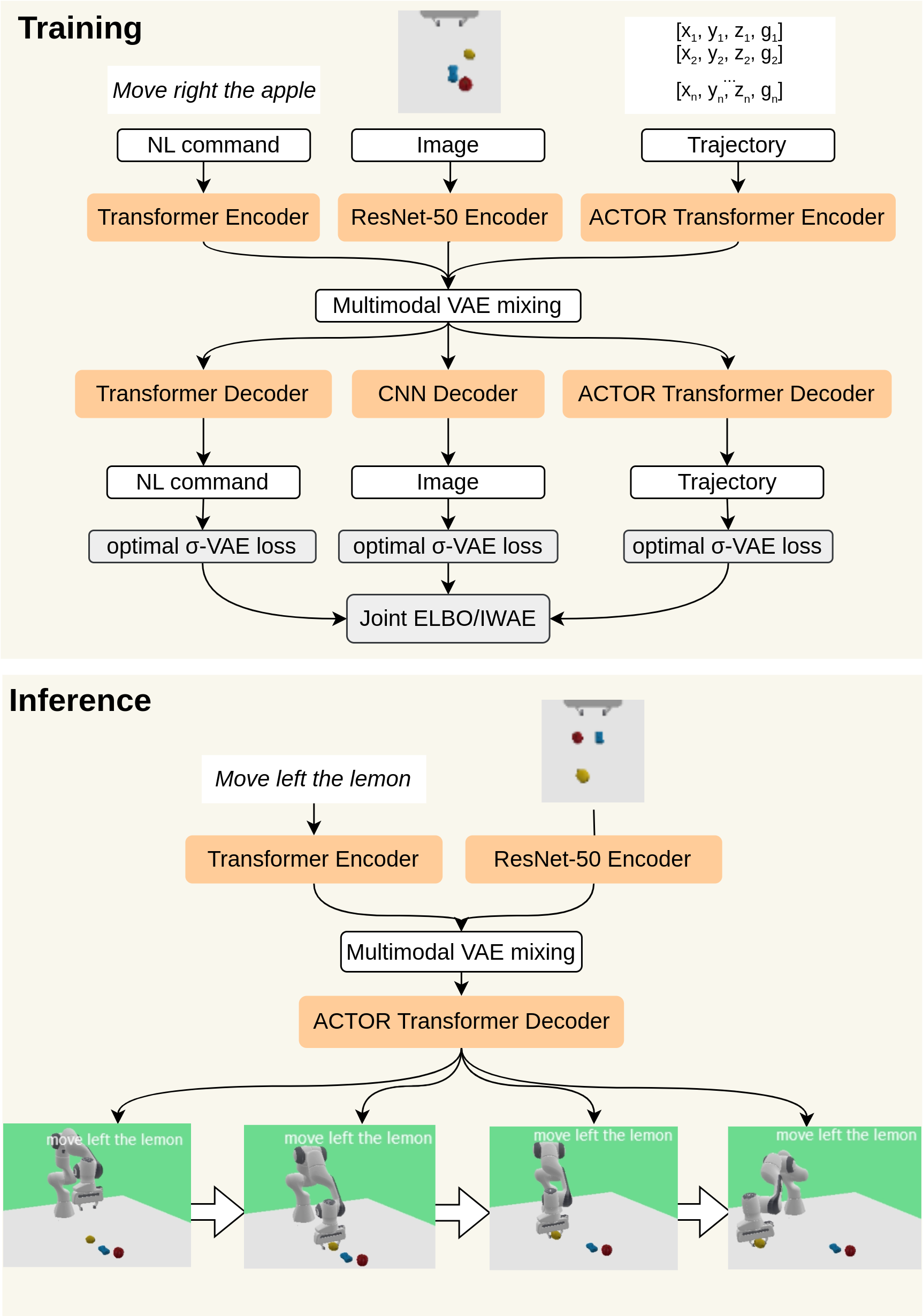}
\caption{The proposed architecture scheme for the training procedure (top) and inference (bottom). During training, the language command, image and motion trajectory are fed into individual model encoders. The multimodal mixing method (MVAE, MMVAE or MoPoE) learns the joint posterior distribution. The reconstruction loss for each modality is calculated using the $\sigma$-VAE loss and used in the ELBO or DReG (for MMVAE) objective. During inference, the model predicts the whole motion trajectory based on the provided image and command.} 
\label{fig:fig1}
\end{figure}

One potentially promising approach towards the vision-language-action fusion challenge would be the use of probabilistic generative models such as Multimodal Variational Autoencoders (VAEs). These models have emerged as a powerful extension of the unimodal Variational Autoencoder and are suitable for mapping multiple modalities into a unified joint space, allowing for the generation of one modality from another \cite{suzuki2022survey}. However, most studies on multimodal VAEs have focused mainly on tasks involving image-to-image or image-to-text mappings, leaving a significant gap in their application to robotic actions \cite{wu2018multimodal}, \cite{shi2019variational}, \cite{lee2021private}, \cite{sutter2021generalized}.

In this paper, we explore the potential of the state-of-the-art (SOTA)  multimodal VAEs in the context of robotic scenarios, where actions are learned from a combination of action demonstrations, images, and natural language instructions (see Fig.~\ref{fig:fig1}). Our primary objective is to understand the challenges and requirements for leveraging VAEs in such a multimodal setting where each modality has a different level of abstraction and complexity (e.g., high-level language instruction versus low-level end-effector trajectory) and there is no additional supervision such as ground truth of the object positions. Our contributions are the following:

\begin{itemize}
    \item We employ three selected SOTA multimodal VAEs and adapt the encoder-decoder architecture for mapping natural language inputs, images and whole motion trajectories in a robotic manipulation task (see Fig.~\ref{fig:fig1}). 
    \item We propose a \deleted{adaptation of the standard VAE architecture} \added{model-independent training objective adjustment that is more suitable for such robotic tasks and improves the performance of the implemented models by up to 55~\% compared to the standard training objectives}.
    \item We train the models on \added{36 synthetic robotic datasets} \added{with variable complexity in terms of the number of tasks, distractors, position variability and task length.} We then evaluate which aspects of the task are the most challenging for the multimodal VAEs.
\end{itemize}

Our project repository is available on GitHub
\footnote{\url{www.github.com/gabinsane/multi-vaes-in-robotics}\label{git}}.


\section{RELATED WORK}
\subsection{Vision-Language-Action mapping in robotics}

In recent years, the most popular approach to multimodal learning of robotic actions based on natural language commands incorporates the usage of pre-trained large language models (LLMs) \cite{wu2023tidybot}, \cite{shah2023lm}, \cite{liang2023code} or large pre-trained vision-language models (VLMs) \cite{zitkovich2023rt}, \cite{driess2023palm}. Such models can parse complex natural language inputs into symbolic task representation and enable semantic reasoning. Moreover, they are often able to generalize to new, unseen data from a small number of training examples. However, they require manual (and often iterative) fine-tuning to condition the model on the specific task, so that the produced results are within the desired range. They also typically cannot be used on a local machine due to the high computational resource requirements. 

In our work, we consider instructions for robotic manipulation which can be learned from scratch and bring our attention to the multimodal mapping among modalities with different levels of abstraction (i.e., symbolic instruction, raw image of the scene and low-level motion trajectory). We employ the generative capability of the SOTA multimodal VAEs to learn such mapping without any supervision. Such an approach is computationally lightweight, does not require manual fine-tuning of the input data and produces more predictable results. Furthermore, multimodal VAEs can also generate natural language captions from the input image and trajectories and can thus serve as an action-recognition module at the same time.

\subsection{Multimodal VAEs}
Multimodal VAEs enable joint integration and reconstruction of two or more modalities. Various approaches to learning joint representation in multimodal VAEs have been proposed \cite{suzuki2016joint}, \cite{wu2018multimodal}, \cite{shi2019variational}, \cite{vasco2020mhvae}, \cite{sutter2021generalized}, \cite{joy2021learning}. The JMVAE model \cite{suzuki2016joint} learns the joint multimodal probability through a joint inference network and requires an additional inference network for each subset of modalities. A more scalable solution is the MVAE model \cite{wu2018multimodal}, where the joint posterior distribution is approximated using the product of experts (PoE). The MMVAE approach \cite{shi2019variational} uses a mixture of experts (MoE) to estimate the joint variational posterior based on individual unimodal posteriors. The MoPoE architecture \cite{sutter2021generalized} combines the benefits of the PoE and MoE approaches by computing the joint posterior for all subsets of modalities. \deleted{Furthermore, the DMVAE model uses shared and modality-specific features in the latent space to learn a disentangled multimodal VAE \cite{lee2021private}. }

Although the above-mentioned models have been evaluated on various image-image or image-text benchmark datasets, they have not yet been tested in a robotic experiment where they would generate whole robotic trajectories. Zambelli et al. \cite{zambelli2020multimodal} proposed a multimodal VAE in which the joint latent space is learned using shared encoder and decoder network layers. Their model was evaluated on a sensomotoric learning scenario with an iCub robot playing a piano. However, the model can predict only one upcoming timestep at a time and thus does not allow learning and distinguishing multiple action classes based on the whole trajectories. Similarly, Meo et al. \cite{meo2021multimodal} adopted the MVAE model \cite{wu2018multimodal} for an adaptive torque controller that estimates the proprioceptive state of a Panda robot from multiple sensory inputs. Such adaptation, however, allows only one-on-one mapping among the modalities (e.g., joint positions to image of the robot) and does not focus on whole complex trajectories and their mapping to language. 

In our research, we explore and propose modifications to state-of-the-art multimodal VAEs, enabling them to infer entire motion trajectories for object manipulation from raw image inputs and natural language instructions in an unsupervised manner. This task presents challenges since each modality provides distinct information at various levels of abstraction. For instance, images convey implicit details about scene objects and their positions, instructions specify target objects and actions symbolically, and the action modality informs about low-level task execution.







\begin{figure*}[!htp]
\centering
\includegraphics[width=1\textwidth]{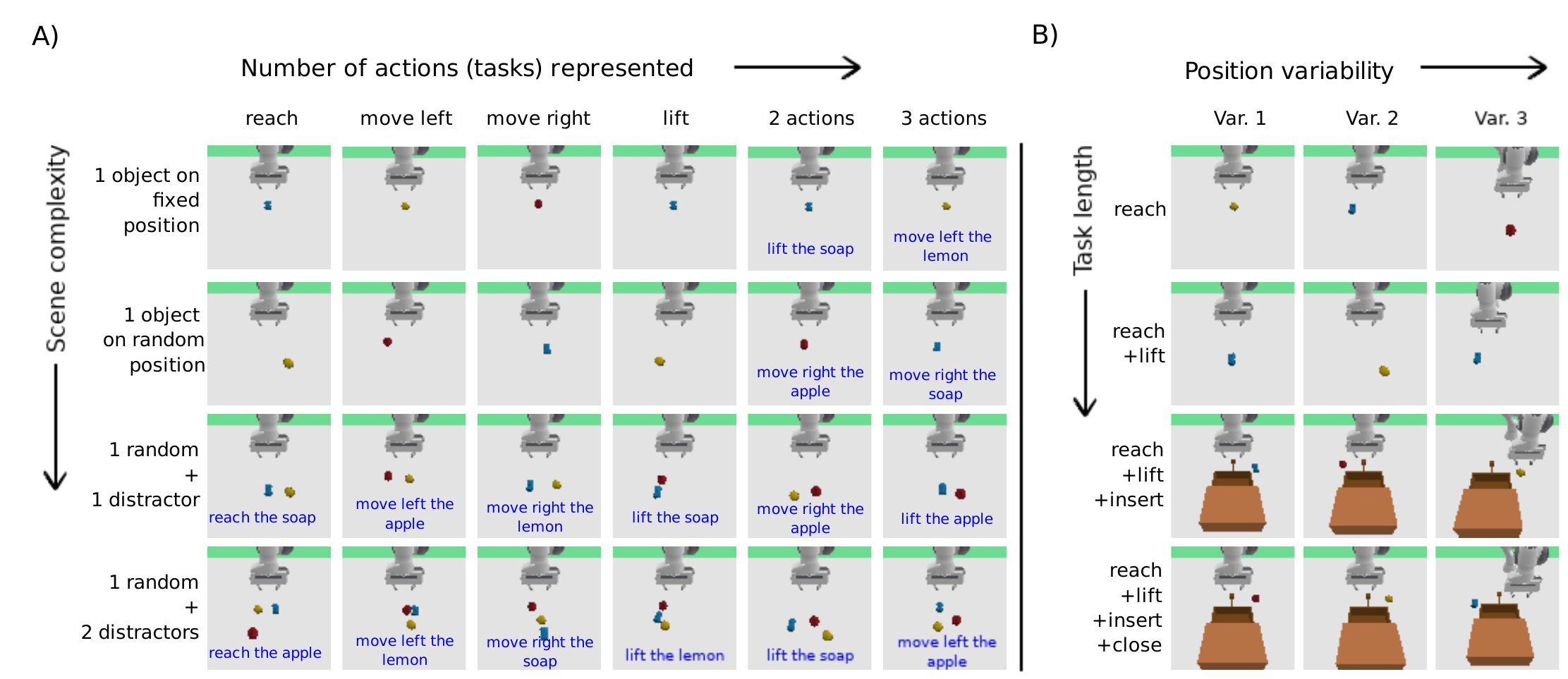}
\caption{Examples from the 36 datasets generated for our experiments. In each dataset, we use the top view of the scene with the robot as the visual input and the instruction as the text input (where necessary for task disambiguation). We also provide the task motion trajectories as another modality. \textbf{A:} the scene complexity (rows) and number of actions (tasks) concurrently represented by the model (columns), e.g., one action, \textit{move right}, is represented by the model, or the model has to represent concurrently more actions (i.e., for 2 actions, the model learns together \textit{move right} and \textit{lift}).  \textbf{B:} the task length (rows) and position variability (columns), i.e., Var. 1 varies object positions along the \textit{x} axis, Var. 2. varies positions along axes \textit{x, y} and Var. 3. additionally varies the robot position. For more details, see Section~\ref{sec:exp}.}
\label{fig:datasets}
\end{figure*}

\section{PRELIMINARIES}
\label{sec:theory}

\subsection{Task Description}

We focus on a robotic manipulation task, where a robotic arm should manipulate randomly positioned objects according to a single top-view RGB image of the scene and a natural language command that specifies the type of action and/or the name of the target object. The robot is controlled using the $x$,$y$ and $z$ end-effector positions and an additional binary value $g$ for opening and closing the gripper. The task requires mapping the actions, images, and language commands into a joint representation so that during inference the model is capable of: 1) generating the action based on the provided NL command and image, and 2) generating the NL description based on the provided trajectory and image. The mapping is achieved by training the model on a synthetic dataset. 

\subsection{VAEs}
\setlength{\abovedisplayskip}{4pt}
\setlength{\belowdisplayskip}{4pt}

In this section, we briefly cover the theoretical background of the state-of-the-art multimodal VAEs that we adapted for the robotic experiment. For more detailed equations, please see the original papers. 

Multimodal VAEs are built on top of the standard (unimodal) VAE as proposed by Kingma et al. \cite{kingma2013auto}, which is a probabilistic encoder-decoder network that learns to approximate the variational posterior $q_\phi(z|x)$ and optimize the evidence lower bound (ELBO):

\begin{equation}
\label{eqn:vae}
\resizebox{0.85\hsize}{!}{$\mathcal{L}_{ELBO}(\boldsymbol{\phi},\boldsymbol{\theta})=\mathbb{E}_{q_{\boldsymbol{\phi}}(\boldsymbol{z}|\boldsymbol{x}_i)}[\log p_{\boldsymbol{\theta}}(\boldsymbol{x}_i|\boldsymbol{z})]- \text{KL}(q_{\boldsymbol{\phi}}(\boldsymbol{z}|\boldsymbol{x}_i) || p(\boldsymbol{z}))$}
\end{equation}


where the first term is the reconstruction loss, i.e., the expected negative log-likelihood of the \textit{i}-th datapoint  $-\sum_{i=1}^{M}\boldsymbol{y}_ilog\hat{\boldsymbol{y}}_i$, 
 and the second term is the Kullback-Leibler divergence between the encoder's distribution $q_\theta(z|x)$ and the prior $p(z)$, which is typically a Gaussian.

In addition to the individual posteriors for each input modality, multimodal VAEs also learn their joint representation, so that it is possible to provide any subset of the modalities on the input at test time. Such a model should then be able to perform cross-generation, i.e., to reconstruct one modality from another. 

In MVAE \cite{wu2018multimodal}, the joint variational posterior of the \textit{N} modalities is approximated using the Product of Experts as 

\begin{equation}
\label{eqn:poe}
\resizebox{.6\hsize}{!}{$q_{\boldsymbol{\phi}}(\boldsymbol{z}|\boldsymbol{x}_1,\ldots,\boldsymbol{x}_N)=\prod_{n} q_{\boldsymbol{\phi}_n}(\boldsymbol{z}|\boldsymbol{x}_n),$}
\end{equation}


assuming that the individual experts are Gaussian and the \textit{N} modalities, $\boldsymbol{x}_1, ..., \boldsymbol{x}_N$ , are conditionally independent
given the common latent variable $z$.

In contrast, MMVAE \cite{shi2019variational} uses a Mixture of Experts to factorize the joint variational posterior as a combination of unimodal posteriors

\begin{equation}
\label{eqn:moe}
\resizebox{0.9\hsize}{!}{$q_{\boldsymbol{\phi}}(\boldsymbol{z}|\boldsymbol{x}_1,...,\boldsymbol{x}_N)=\sum_n\alpha_n\cdot q_{\boldsymbol{\phi}_n}(\boldsymbol{z}|\boldsymbol{x}_n)$, where $\alpha_n=1/N$},
\end{equation}

assuming that all inputs are of comparable complexity and all of them are present during the whole training. Although the MMVAE model can be trained with the traditional ELBO objective, the authors propose using the importance-weighted autoencoder (IWAE) objective together with the reparametrised gradient estimator DReG \cite{shi2019variational}. We thus use IWAE with DReG for the MMVAE model in our experiments.

The Mixture of Products of Experts (MoPoE) model \cite{sutter2021generalized} is then a generalization of the MVAE and MMVAE models: 

\begin{equation}
\label{eqn:mopoe}
q_{MoPoE}(\mathbf{z} \mid X)=\frac{1}{2^N} \sum_{X_k \in \mathcal{P}(X)} q_{PoE}\left(\mathbf{z} \mid X_k\right),
\end{equation}

where $q_{PoE}\left(\mathbf{z} \mid X_k\right) \propto \prod_{\mathbf{x}_n \in X_k} q_{\phi_n}\left(\mathbf{z} \mid \mathbf{x}_n\right)$ and $\mathcal{P}(X)$ is the powerset of the $N$ modalities. For more details, see the original paper \cite{suzuki2022survey}. 

\deleted{The DMVAE model \cite{lee2021private} factorises the latent space into modality-specific (private) latents $\boldsymbol{z}_{p1,..,pN}$ and shared latents $\boldsymbol{z}_{s1,..,sN}$. For the shared latent representation, the model learns to achieve $\boldsymbol{z}_{s1}$ = $\boldsymbol{z}_{s2}$ = \ldots = $\boldsymbol{z}_{sN}$ by employing the PoE approach (see Eq.~\ref{eqn:poe}). For more details, please see the original paper  \cite{lee2021private}. }

\section{Multimodal VAE for robotic scenarios}
\label{sec:architecture}
First, we adjust the multimodal VAE-based architecture so that it can integrate the three input modalities in the vision-language-action scenario and produce motion trajectories from images and commands during inference. This adaptation is independent of the multimodal mixing methods (described in Sec.~\ref{sec:theory}, see also Fig.~\ref{fig:fig1}). \added{Second, we employ the $\sigma$-VAE reconstruction loss term \cite{rybkin2021simple} in a multimodal scenario and compare with the classical mean squared error loss. }

\begin{figure}[!htp]
\centering
\includegraphics[width=.7\columnwidth]{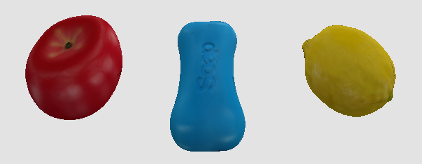}
\caption{The objects used in our experimental setup: apple, soap and lemon.} 
\label{fig:objects}
\end{figure}
\vspace{-2mm}

\subsection{Architecture}

In this section, we describe the encoder and decoder networks that were used for individual input modalities. You can see the overall architecture in Fig.~\ref{fig:fig1}.

\subsubsection{Images}
We fed a single RGB image with a top view of the scene to the VAE encoder. To extract the underlying image features, we used a ResNet-50 network pre-trained on ImageNet \cite{he2016deep} as the encoder and 3 fully connected layers followed by 3 convolutional layers as the decoder.

\subsubsection{Robotic Actions}
\label{ssec:actor}
We adapted the encoder and decoder networks for the motion trajectory of the ACTOR model designed for the generation of human motion, conditioned on the action label and the sequence length \cite{petrovich2021action}. ACTOR is a VAE with transformer encoder and decoder networks with positional encodings, allowing variable-length sequence generation. To allow conditioning, the authors used a learnable bias $b_a^{token}$ to shift the latent representation to an action-dependent space. We do not use this feature as the latent representation is already shifted after mixing with language instructions and the image in the joint posterior. We use an 8-layer (each with self-attention and feed-forward network) transformer network with 2 heads and 1024 hidden dimensions as the encoder and an 8-layer (each with self-attention, multi-head attention and feed-forward network) transformer network with 2 heads and 1024 hidden dimensions as the decoder. The model can handle the generation of variable-length trajectories due to positional encodings.

\subsubsection{Language instructions}
For the language encoder and decoder, we used transformer networks similar to the ones we used for the actions (see Sec.~\ref{ssec:actor}). The main difference is that we added a trainable embedding layer to process the text one-hot encodings on the input and compressed them into 2-dimensional features. These features were then fed into the positional encoder followed by the transformer. The decoder architecture remained the same as for the actions. Compared to the action encoder and decoder networks, we used different parameters - 1 layer, 2 heads, and 128 hidden dimensions for both the encoder and decoder.\looseness=-1

\subsection{Proposed training adjustment}
\label{ssec:reconloss}
The reconstruction loss term used in the ELBO objective can have an impact on the final reconstruction quality, as well as the stability of the model performance \cite{yan2021videogpt}. 
Compared to the standardly used mean squared error (MSE) or negative log-likelihood (NLL), we find that the quality of the produced reconstructions of all multimodal models is improved when using the \textit{optimal} $\sigma$-VAE proposed by Rybkin et al. \cite{rybkin2021simple} which uses an analytical estimate of the decoder variance:

\setlength{\abovedisplayskip}{5pt}
\setlength{\belowdisplayskip}{5pt}
\begin{equation}
\label{eqn:sigma}
\sigma^{* 2}=\underset{\sigma^2}{\arg \max } \mathcal{N}\left(x \mid \mu, \sigma^2 I\right)=\operatorname{MSE}(x, \mu),
\end{equation}

where $\operatorname{MSE}(x, \mu)=\frac{1}{D} \sum_i\left(x_i-\mu_i\right)^2$. Here, the maximum likelihood estimate of the variance $\sigma$ given a known mean $\mu$ is the average squared distance from the mean. \deleted{Based on our comparative study (see Sec.~\ref{ssec:resultslosses}), we use the $\sigma$-VAE objective for all of our experiments} \added{In our experiments in Sec.~\ref{sec:exp}, we compare the MSE loss with the $\sigma$-VAE objective on 36 robotic datasets. We do not compare with the NLL loss as the results produced by training with this reconstruction loss are significantly worse than MSE or $\sigma$-VAE.}



\section{Experimental Setup}
\label{sec:exp}
\subsection{Simulated robotic environment}
\label{ssec:env}
We used an adapted version of the LANRO (Language Robotics) simulator for our experiments \cite{roder2022language}. The environment includes the Franka Emika robot equipped with a 2-finger gripper pointing downward, controlled using end-effector positions and binary values for gripper closing or opening. The robot is positioned on top of a table where objects are spawned during the experiment. The top-view camera is located above the centre of the table. For dataset collection, we used \added{3 textured objects -  a soap, an apple and a lemon} (see Fig.~\ref{fig:objects}). In each trial, one or more objects are spawned 60-80 cm from the robot base (see Subsections~\ref{ssec:scenecom} and \ref{ssec:posvar}) and the environment provides a natural language instruction that disambiguates the task, i.e., provides information on the desired action in multi-task scenarios (\textit{reach, }\textit{lift}, \textit{move right}, \textit{move left}) and specifies the object in the case of multi-object datasets (e.g., \textit{move left the soap}). \deleted{To make the instructions more variable, we also randomly use synonyms for the actions - \textit{drag} and \textit{shift} as synonyms for \textit{move}, and \textit{hoist} and \textit{raise} as a synonym for \textit{lift}} \added{In an additional set of experiments, we also include tasks where the robot has to place an object into a drawer and close it}. Examples of setup and specific tasks are shown in Fig.~\ref{fig:datasets}.\looseness=-1

\subsection{Scene complexity and task types}
\label{ssec:scenecom}

To explore the multimodal VAE capabilities, we created data with task demonstrations performed by the robot based on the provided image of the scene and language instructions (used for task disambiguation when necessary). Using the virtual setup described in Section~\ref{ssec:env}, we first generated six datasets, each requiring the representation of a different number and type of actions (tasks), with four distinct levels of scene complexity for each case. This resulted in a total of 24 (6x4) datasets  (see Fig.~\ref{fig:datasets}A). \added{They differ based on the \textbf{scene complexity}} as follows:

\begin{itemize}
    \item \textbf{1 object in fixed position} - in each trial, one randomly selected object out of 3 is placed in the scene, always in the same location 60 cm from the robot base. The robot then performs one of the actions on the object (e.g., \textit{move left}).
    \item \textbf{1 object in a random position} -  one randomly selected object out of 3 is placed in the scene in a random position varying between 60-80 cm from the robot in the \textit{x}-axis and between -10 and +10 cm in the y axis.
    \item \added{\textbf{1 object in a random place + 1 distractor} - one randomly selected object is placed in the scene in a random location. Additionally, another (different) object is placed somewhere in the scene. The instruction then specifies which object should be moved.}
    \item \added{\textbf{1 object in a random place + 2 distractors} - similar to the previous scenario, but with two distractors in the scene.}
\end{itemize}

\added{For each of the dataset scenes above, we generated 6 different training sets comprising a various number and types of tasks (robot actions) to perform on the selected object. The variability is the following: }

\begin{itemize}
    \item \added{\textbf{reach} - the robot has to approach the selected object within the distance of 6 cm from the centre of the object}, which is the optimal position for grasping
    \item \added{\textbf{move left} - the robot has to lift the object and move it to the left by at least 10~cm.}
    \item \added{\textbf{move right} -  the robot has to lift the object and move it to the right by at least 10~cm.}
    \item \added{\textbf{lift} -  the robot has to lift the object up by at least 10~cm.}
    \item \textbf{2 actions (move right or lift)} -  the robot has to either perform action \textit{lift} or \textit{move right} based on the provided instruction.
    \item \textbf{2 actions (move right or move left or lift)} -  the robot has to either perform action \textit{lift} or \textit{move right} 
 or \textit{move left} based on the provided instruction.
\end{itemize}


All of the aforementioned trainsets comprised 10,000 samples in single-task scenarios and 20,000 and 30,000 samples in the 2-action and 3-action datasets, respectively. Each trial involved collecting data from 3 modalities: the initial scene image (RGB, size 64x64x3), natural language instructions (one-hot encodings for each word), and the robotic action trajectory (4 values per timestep: $x$, $y$, and $z$ coordinates of the end effector, and a binary value for the gripper).

\subsection{Position variability and task length}
\label{ssec:posvar}

\begin{figure}[!htp]
\centering
\includegraphics[width=.99\columnwidth]{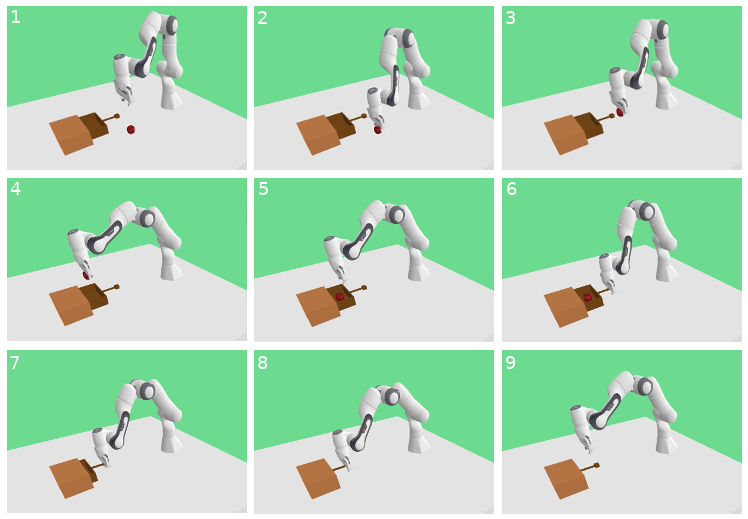}
\caption{Example of the \textit{reach+lift+insert+close} sequence (ordered by numbers), which is the longest task in our datasets. The length of the sequence is up to 68 timesteps. For more examples, see the attached video.} 
\label{fig:sequence}
\end{figure}

\added{Next, to test the robustness of the multimodal VAEs towards the position variability and task length, we generated datasets with four different task lengths (action sequences), each featuring three levels of positional variability. This resulted in 12 datasets (see Fig.~\ref{fig:datasets}B). The tasks used here were the following:}
\begin{itemize}
    \item \textbf{reach} - the robot has to approach the selected object under the distance of 6 cm (corresponds to \textit{reach} in Subsection~\ref{ssec:scenecom})
    \item \textbf{reach+lift} - the robot has to lift the object and move it to the left by at least 10~cm (corresponds to \textit{lift} in Subsection~\ref{ssec:scenecom}).
    \item \textbf{reach+lift+insert} -  the robot has to lift the object and place it inside an open drawer.
    \item \textbf{reach+lift+insert+close} -  the robot has to lift the object, place it into the drawer and close it (see example in Fig.~\ref{fig:sequence}).
\end{itemize}

For each of the above tasks, we generated 3 trainsets differing in the amount of positional variability:
\begin{itemize}
    \item \textbf{Variability 1} - the object (and drawer) position varied only along the \textit{x} axis (e.g., closer or further from the robot) within the range of 20~cm.
    \item \textbf{Variability 2} - the positions of objects (and drawers) varied along the \textit{x} and \textit{y} axes by 20~cm (for \textit{reach} and \textit{reach+lift}, the variability of the position is the same as in the datasets used in Section~\ref{ssec:scenecom}).
    \item \textbf{Variability 3} - the same as Variability 2, but the robot base position also varied along the \textit{y} axis (e.g., along the table edge) by 40~cm.
\end{itemize}

We chose a positional variability of 20 cm in these experiments to ensure that the object remains visible in the camera image and to avoid interference with the drawer. As the task was always the same within each dataset (and there were no distractors), we did not include the natural language instructions in these datasets and provided only the image and the robotic action trajectories to the models.  The number of time steps varied between tasks and trials, ranging from 13 time steps (\textit{reach}) to 68 timesteps (\textit{reach+lift+insert+close}).



\subsection{Training setup}

We compared the MVAE, MMVAE, and MoPoE models using the identical architecture proposed in Section~\ref{sec:architecture} and visualized in Fig.~\ref{fig:fig1}.  The compared models thus have the same encoder and decoder networks, allowing for a direct comparison of the multimodal VAE approaches, with differences primarily arising from the mixing approach. We trained each model for 400 epochs and with the optimal latent space size based on a hyperparameter grid search (the final training parameters can be found on our GitHub \textsuperscript{\ref{git}}). After training, we used the test data to create previously unseen scenes in the LANRO simulator and used the model to predict the motion trajectories for the robot. 




\section{RESULTS AND DISCUSSION}

\begin{figure}[!htp]
\centering
\includegraphics[width=.99\columnwidth]{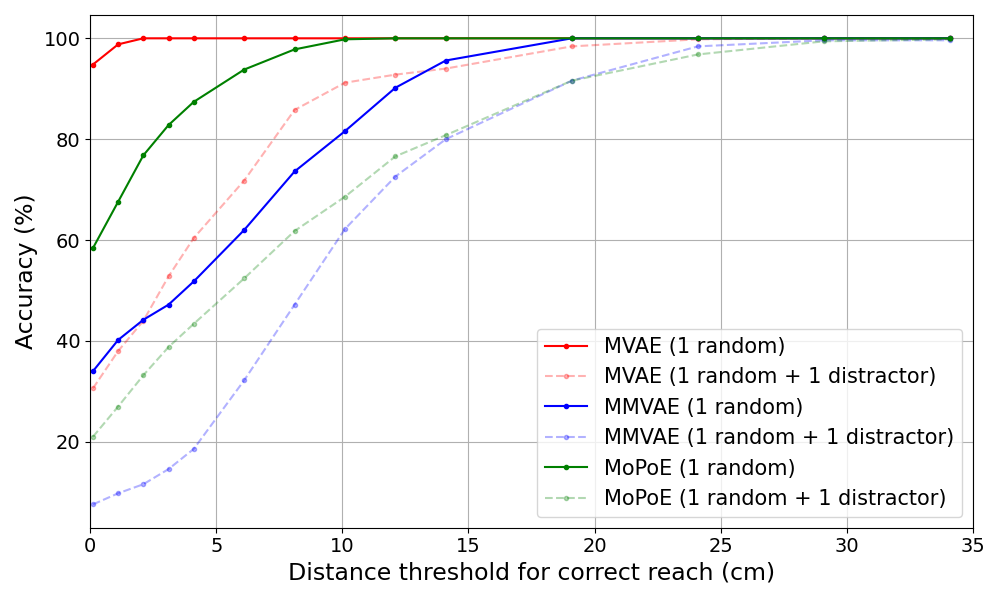}
\caption{Accuracy for the reach task based on the threshold of the final distance between the gripper and target reach position, which is 6 cm from the object centroid (optimal distance for grasping). We show the results for MVAE, MMVAE and MoPoE. \textit{1 random} corresponds to the dataset with 1 randomly placed object, the data for \textit{1 random + 1 distractor} also included a random distractor in the scene.} 
\label{fig:thresholds}
\vspace{-4mm}
\end{figure}

\begin{table*}[htp!]
\small
\centering
\caption{Task accuracy for the three multimodal VAE models based on the scene complexity (rows) and represented actions (columns), i.e., how many and which actions were in the training data (\textit{2 actions}: \textit{lift} and \textit{move right}, \textit{3 actions} also included \textit{move left}). We compare results for training with the mean squared error (MSE) reconstruction loss and proposed $\sigma$-VAE adaptation.}
\label{table:results}
\renewcommand{\arraystretch}{1.2}
\resizebox{0.99\textwidth}{!}{%
\fontsize{6}{7}\selectfont 
\begin{tabular}{cc|c|cc|cc|cc|cc|cc|cc}  
\multicolumn{3}{c}{}  &  \multicolumn{12}{c}{\cellcolor{Gray}\textbf{Represented actions (tasks)}}  \\
 \\[-0.7em]
 \multicolumn{2}{c|}{\multirow{3}{*}{\centering Accuracy [\%]}} & \multirow{3}{*}{\centering Model}  & \multicolumn{2}{c|}{\textbf{Reach}} & \multicolumn{2}{c|}{\textbf{Move left}} & \multicolumn{2}{c|}{\textbf{Move right}} & \multicolumn{2}{c|}{\textbf{Lift}}  & \multicolumn{2}{c|}{\textbf{2 Actions}} & \multicolumn{2}{c}{\textbf{3 Actions}}  \\
 & &  &  \multirow{2}{*}{MSE}  &  \vtop{\hbox{\strut \centering $\sigma$-VAE}\hbox{\strut (proposed)}}  &  \multirow{2}{*}{MSE} &  \vtop{\hbox{\strut \centering $\sigma$-VAE}\hbox{\strut (proposed)}}   &  \multirow{2}{*}{MSE}  & \vtop{\hbox{\strut \centering $\sigma$-VAE}\hbox{\strut (proposed)}}  &  \multirow{2}{*}{MSE}  &  \vtop{\hbox{\strut \centering $\sigma$-VAE}\hbox{\strut (proposed)}}  &  \multirow{2}{*}{MSE}  & \vtop{\hbox{\strut \centering $\sigma$-VAE}\hbox{\strut (proposed)}}  &  \multirow{2}{*}{MSE}  & \vtop{\hbox{\strut \centering $\sigma$-VAE}\hbox{\strut (proposed)}}   \\
\hline
\parbox[t]{2mm}{\multirow{12}{*}{\rotatebox[origin=c]{90}{\textbf{Scene complexity}}}}  &  \multirow{3}{2cm}{1 fixed} & MVAE & 100 & 100& 100&100 &100 &100 & 100& 100& 59 & \textbf{67} & 28 & \textbf{30} \\& & MMVAE & 100 &  100 & 100 & 100 & 100 & 100 & 100 & 100 & 100 & 100 & 67 & \textbf{100} \\
& & MoPoE & 100 & 100 &100 & 100 &100 & 100&100 &100 &66 & \textbf{67} &12 & \textbf{22} \\
\cline{2-15}
&\multirow{3}{2cm}{1 random} & MVAE & 67 & \textbf{97} & 50 & \textbf{67} & 51 & \textbf{63} & 39 & \textbf{54} & 22 & \textbf{28} & 18 & \textbf{20} \\
& & MMVAE & 7 & \textbf{35} & 5 & \textbf{13} & 1 & \textbf{5} & 4 & \textbf{9} & 4 & 4 & 1 & \textbf{6} \\
& & MoPoE & 5 & \textbf{60} & 11 & 11 &  8&\textbf{17} &2 &\textbf{10} &4 & \textbf{15} & 10 & \textbf{21} \\
\cline{2-15}
& \multirow{3}{2cm}{1 random \newline +1~distractor} & MVAE & 21 & \textbf{32} & 19 & \textbf{21} & 21 & \textbf{23} & 10 & \textbf{14} & 16 & \textbf{19}  & 12 & 12 \\
& & MMVAE & 5 & \textbf{7} & 2 & \textbf{6} & 6 & 6 & 3 & \textbf{4} & 3 & \textbf{4} & 1 & \textbf{3} \\
& & MoPoE & 10 & \textbf{21} & 10 & \textbf{12} & 8& \textbf{15}& 2&\textbf{4} &4 & \textbf{6}&6 & 6 \\
\cline{2-15}
& \multirow{3}{2cm}{1 random \newline +2 distractors} & MVAE & 15 & \textbf{20} & 15 & \textbf{16} & 11 & \textbf{14} & 4 & \textbf{6} & 10 & 10 & 6 & \textbf{7} \\
& & MMVAE & 0 & \textbf{6} & 0 & \textbf{5} & 1 & \textbf{2} & 0 & \textbf{1} & 1 & \textbf{4}& 4 & \textbf{5} \\
& & MoPoE & 8 & \textbf{9} & 8 & \textbf{10}& 6 & \textbf{10} &1 & \textbf{3}&3 & \textbf{7}& 3& 3 \\
\end{tabular}}
\end{table*}

After training, we tested the multimodal VAEs in our test sets consisting of 500 trials. We provided only the image of the scene and natural language command (when needed for disambiguation) and the model had to generate the correct action. We considered the generated sample correct only if it reached or moved the object within the thresholds used to generate the training data (i.e., the gripper approached the object within less than 6 cm distance for the \textit{reach} task, measured from the centroid of the object). \vspace{-1mm}

\subsection{Comparing the reconstruction loss terms}

\added{In Table~\ref{table:results}, we compare the mean squared error loss (MSE) used as the reconstruction loss term with our proposed adjustment, the $\sigma$-VAE loss \cite{rybkin2021simple} (see Section~\ref{ssec:reconloss}). For all models and all datasets (shown in Fig.~\ref{fig:datasets}A), $\sigma$-VAE performed the same or better than MSE. The total average improvement in accuracy over all models and scenarios is 6.4~\% (min. 0~\%, max. 55~\%). These results indicate that the batch-wise analytic estimate of the variance is beneficial also for multimodal VAE models, regardless of the multimodal mixing method.}

\subsection{Impact of the scene complexity and number of tasks}

\added{We find that all multimodal VAEs are able to learn a single action in a scenario with a non-variable setup with one object on a fixed position (see Table~\ref{table:results}, first row, columns 1-8). Although training two or three actions at the same time causes a continuous decrease in accuracy for the MVAE and MoPoE models, MMVAE was not affected by the number of tasks (see Table~\ref{table:results}, row 1, columns 9-12). The \textit{reach} task accuracy provides insights into the model's capability to localize the correct object based solely on vision. Given the accuracy differences between \textit{reach} and individual actions such as \textit{move left} are generally not substantial, we can assume that the most challenging aspect lies indeed in the mapping between the pixel space into cartesian positions.}

\added{As for the scene complexity, all models were significantly impacted by the variability in object positions and by the presence of distractors. In Fig.~\ref{fig:thresholds}, we show the accuracy of the \textit{reach} task based on the threshold of the distance from the target object. When using one randomly positioned object, all of the inferred trajectories for MVAE and MoPoE were within the maximum distance of 10 cm from the target reach position (for MMVAE, the maximum distortion was 19 cm). However, when introducing one distractor, the proportion of severely distorted trajectories (more than 15 cm) increased. Such results indicate that mapping between pixel and cartesian space remains a challenging task for the SOTA multimodal VAEs. The incorporation of additional vision pre-processing modules might be beneficial.}

\subsection{Impact of the position variability and task length}

\added{In Table~\ref{table:results2}, we show how task accuracy is influenced by both task length and position variability.  Although there was a slight decrease in accuracy due to variations in object and robot positions, this impact was not as significant as the effect of the task length. Notably, both MVAE and MMVAE faced challenges in executing the longest task, which involved placing an object into the drawer and closing it. A potential solution to improve performance in such long-horizon tasks could involve training individual VAE experts for each subtask, taking into account variable object and robot positions, and then chaining these experts together to execute the entire task seamlessly.}

\subsection{Comparing the multimodal VAE models}

\added{Tables~\ref{table:results} and \ref{table:results2} show that the MVAE model \added{consistently} outperforms both MMVAE and MoPoE \deleted{in}\added{across} most tasks. MVAE achieved \>50~\% accuracy in all single-task datasets with one randomly placed object. On the other hand, MMVAE demonstrates higher accuracy on multi-task datasets (learning 2 or 3 actions simultaneously), yet only in the scenario with one fixed object, indicating limited robustness towards scene complexity. MoPoE, as a combination of both MVAE and MMVAE approaches, produced results with accuracies between the two models.}


\begin{table}[htp!]
\small
\centering
\caption{Inference accuracies (\%) for MVAE and MMVAE evaluated on 4 tasks: reach, reach + lift, reach + lift + insert into drawer, reach + lift + insert + close drawer. In dataset Var. 1, the objects' positions varied only along the \textit{x} axis, in Var.~2, the objects moved along both the \textit{x} and \textit{y} axes and in dataset Var.~3, we also varied the position of the robot. We used the $\sigma$-VAE objective to train both models.}

\resizebox{0.99\columnwidth}{!}{%
\fontsize{5}{6}\selectfont 
\begin{tabular}{l|c|c|c|c} 

Accuracy [\%] & Model & Var. 1 & Var. 2 & Var. 3  \\
\hline
\multirow{2}{*}{Reach} & MVAE & \textbf{100}  & \textbf{97}  &  \textbf{92} \\
 & MMVAE & 69  & 67  & 64  \\
\hline
\multirow{2}{*}{Reach+Lift} & MVAE & \textbf{72}  &  \textbf{54} &  \textbf{50} \\
 & MMVAE & 52  & 39  &  25 \\
\hline
\multirow{2}{*}{Reach+Lift+Insert} & MVAE &  \textbf{58} &  \textbf{45} &  \textbf{37} \\
 & MMVAE & 39  &  12 &  11 \\
\hline
\multirow{2}{*}{Reach+Lift+Insert+Close} & MVAE &  \textbf{16} & \textbf{14}  &  \textbf{13} \\
 & MMVAE &  1 & 0  & 0  \\
\end{tabular}}
\label{table:results2}
\end{table}

\vspace{-4mm}
\section{CONCLUSIONS}
We employed three SOTA multimodal VAE models for vision-language-action mapping in robotic kinesthetic teaching \added{and performed their systematic evaluation on 36 simulated datasets}. We proposed an adapted architecture (see Fig.~\ref{fig:fig1}) with fine-tuned encoder and decoder networks for each modality and employed the $\sigma$-VAE objective which improved the models' performance by up to 55~\%. We showed that the most suitable is the MVAE model, which \added{had the most robust performance in terms of task length and scene complexity or position variability}. 
In our future work, we would like to further explore the models' capabilities to recognize the correct object in a multi-object scene, as this task showed to be out of scope for the SOTA multimodal VAEs when combined with action learning.  





\section*{ACKNOWLEDGMENT}

\small
This work was supported by the Czech Science Foundation (GA {\v C}R) grants no. 21-31000S and 23-04080L, by the Ministry of Education, Youth and Sports of the Czech Republic through the e-INFRA CZ (ID:90254) and co-funded by the European Union under the project Robotics and Advanced Industrial Production (reg. no. CZ.02.01.01/00/22\_008/0004590). 


\bibliographystyle{IEEEtran}
\bibliography{bibliography}

\end{document}